\pdfoutput=1

\documentclass[11pt]{article}

\usepackage[final]{acl}

\usepackage{times}
\usepackage{latexsym}
\usepackage{adjustbox}
\usepackage{enumitem}
\usepackage{booktabs}
\usepackage[T1]{fontenc}
\usepackage{natbib}
\usepackage[utf8]{inputenc}

\usepackage{microtype}
\usepackage{multicol}
\usepackage{multirow}
\usepackage{multirow}
\usepackage{makecell}
\usepackage{array}
\usepackage{booktabs}
\usepackage{algorithm}
\usepackage{algpseudocode}
\usepackage{amsmath}
\usepackage{colortbl}
\usepackage{hhline}
\usepackage{inconsolata}
\usepackage{marginnote}
\usepackage{graphicx}
\usepackage{amssymb}
\usepackage{pifont}
\usepackage[flushmargin]{footmisc} 
\title{Dual-Cluster Memory Agent: Resolving Multi-Paradigm Ambiguity in Optimization Problem Solving}

\author{Xinyu Zhang$^{1,2}$\thanks{These authors contributed equally to this work.},\; Yuchen Wan$^{1,2\ast}$,\; Boxuan Zhang$^{1,2}$,\; Zesheng Yang$^{1,2}$,\;  \\ 
\textbf{Lingling Zhang}$^{1,2}$\thanks{Corresponding author}, \; \textbf{Bifan Wei}$^{1,2}$,\;
\textbf{Jun Liu}$^{1,3}$\\
{$^{1}$School of Computer Science and Technology, Xi’an Jiaotong University}\; \\
{$^{2}$Ministry of Education Key Laboratory of Intelligent Networks and Network Security, China} \; \\
{$^{3}$Shaanxi Province Key Laboratory of Big Data Knowledge Engineering, China} \; \\
\texttt{{zhang1393869716}@stu.xjtu.edu.cn, \{zhanglling,liukeen\}@xjtu.edu.cn}
}

\begin{document}
\maketitle

\begin{abstract}
Large Language Models (LLMs) often struggle with structural ambiguity in optimization problems, where a single problem admits multiple related but conflicting modeling paradigms, hindering effective solution generation.
To address this, we propose \textbf{Dual-Cluster Memory Agent (DCM-Agent)} to enhance performance by leveraging historical solutions in a training-free manner.
Central to this is Dual-Cluster Memory Construction. 
This agent assigns historical solutions to modeling and coding clusters, then distills each cluster's content into three structured types: \textit{Approach}, \textit{Checklist}, and \textit{Pitfall}. 
This process derives generalizable guidance knowledge.
Furthermore, this agent introduces \textbf{Memory-augmented Inference} to dynamically navigate solution paths, detect and repair errors, and adaptively switch reasoning paths with structured knowledge.
The experiments across seven optimization benchmarks demonstrate that DCM-Agent achieves an average performance improvement of 11\%- 21\%.
Notably, our analysis reveals a ``knowledge inheritance'' phenomenon: memory constructed by larger models can guide smaller models toward superior performance, highlighting the framework's scalability and efficiency.
\end{abstract}

\section{Introduction}
Optimization problems underpin operations research, supporting applications from supply chain logistics to economic forecasting \cite{liu2025mm, liu2023milp, belil2018milp}.
Traditionally, solving these problems requires an intensive process \cite{meerschaert2013mathematical}, where domain experts manually translate textual descriptions into formulations.
Recently, Large Language Models (LLMs) have demonstrated remarkable reasoning capabilities across diverse domains, including science reasoning \cite{zhang-etal-2025-physreason, zhang2025diagram, zhang2025memory}, cognitive reasoning \cite{zhang2025cognitive}, and  temporal analysis~\cite{zhang2022tn}. These advances have also reshaped optimization modeling, leveraging domain knowledge to automate the conversion of textual descriptions into optimization models, thereby mitigating dependency on human expertise \cite{sinhaautoopt, zhaostructure}.
However, harnessing LLMs for this purpose remains a challenge, leading to diverse preliminary explorations \cite{wang-etal-2025-ormind, jiangllmopt, chen2025improving}.
\par
\par
\begin{figure}[t]
\centering
\includegraphics[width=0.47\textwidth]{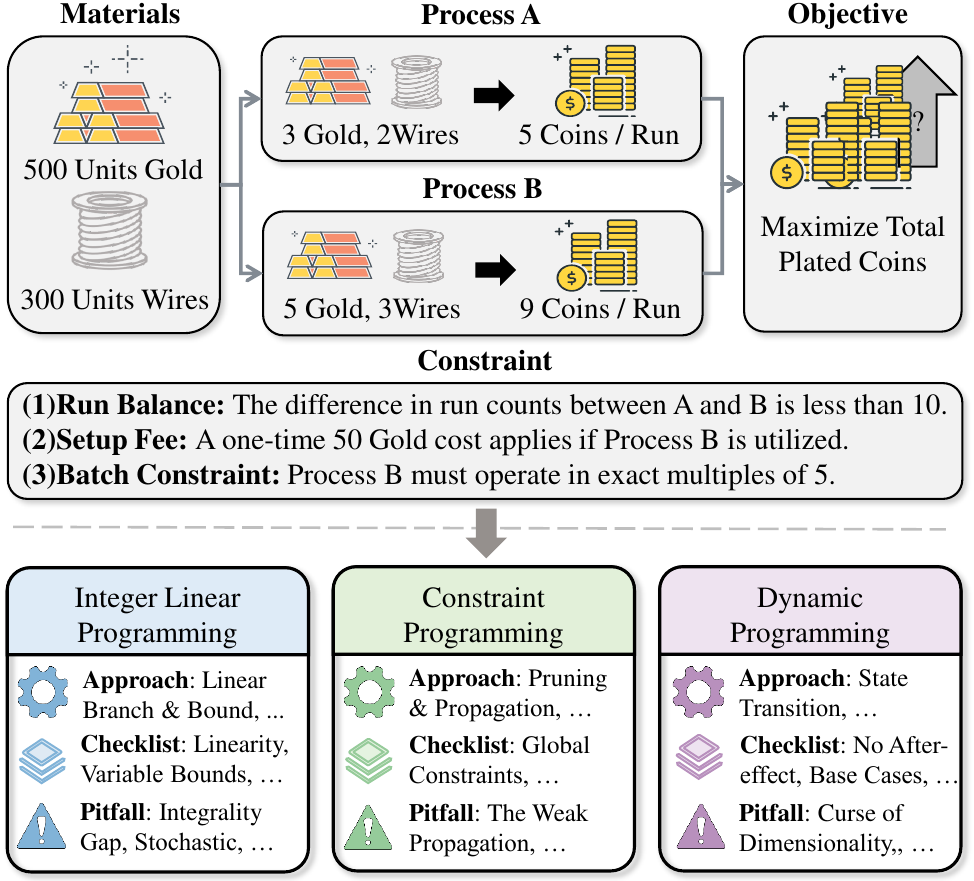}
\vspace{-3pt}
\caption{Illustration of a single production planning problem formalized via distinct algorithmic paradigms.}
\label{fig:intro}
\vspace{-14pt}
\end{figure}

A fundamental bottleneck impeding current methods is cognitive interference from entangled modeling paradigms.
As shown in Figure \ref{fig:intro}, this presents conflicting signals, where logical constraints (e.g., exact multiples) suggest Constraint Programming (CP), resource maximization implies Integer Linear Programming (ILP), and sequential stage dependencies signal Dynamic Programming (DP).
While these features are relevant, their simultaneous presence acts as a confounding distractor.
\par
\par
This complexity exposes deficiencies in current paradigms.
Fine-tuning approaches are misled by this interference, mechanically applying memorized templates when subtle variations necessitate alternative strategies \cite{wang2025large}.
Agentic frameworks struggle with these trade-offs during verification; facing ambiguous algorithmic choices, they rely on the static prompts (e.g., ``Check if correct'') that lack the granularity to detect paradigm-specific pitfalls \cite{huanglarge}, such as distinguishing between linearity gaps in ILP and recurrence validity in DP \citep{zhang2025or}.
This reveals a fundamental tension: the interference problem demands flexibility to navigate paradigm ambiguity, yet also requires targeted knowledge to verify paradigm-specific correctness.
\par
To address this, we propose the \textbf{Dual-Cluster Memory Agent (DCM-Agent)}, a training-free framework that balances flexibility and structure by externalizing reasoning patterns into historical archives.
Instead of relying on parameter updates, this framework decouples abstract modeling from precise coding by directly capitalizing on historical solution archives.
Central to this framework is the \textbf{Dual-Cluster Construction}, which resolves structural ambiguity by organizing historical data, ranging from canonical successes to persistent failures, into a bipartite graph that quantifies the relationships between the independent \textit{Modeling Cluster} and the \textit{Coding Cluster}.
This process distills the raw experience nodes within each cluster into three cluster-level structured knowledge tiers: \textit{Approach}, \textit{Checklist}, and \textit{Pitfall}.
By explicitly mapping the decision space, this design enables the agent to filter out interfering noise and translate isolated solutions into robust structural knowledge.
\par
Building upon the dual-cluster memory, we introduce \textbf{Memory-Augmented Inference}.
Unlike conventional static prompting, this mechanism effectively solves new problems by retrieving relevant clusters that offer corresponding structured knowledge to guide.
The inference is executed through an iterative generate-verify-repair-backtrack pipeline that operates dynamically.
By leveraging the retrieved \textit{Pitfalls} and \textit{Checklists}, this pipeline systematically steers the generation process, allowing the agent to detect method-specific errors (e.g., integrality gaps versus recurrence failures) and autonomously backtrack to alternative reasoning paths when a chosen paradigm proves infeasible.
\par
We evaluate DCM-Agent across seven diverse optimization benchmarks, where it achieves an average performance improvement of 11\%--21\% compared to standalone LLMs. 
The results demonstrate that DCM-Agent maintains consistent state-of-the-art accuracy across various model scales without the computational overhead of training, offering a superior trade-off between solution precision and efficiency. 
Crucially, our analysis reveals a ``knowledge inheritance'' phenomenon: memory constructed by larger models can be effectively transferred to guide smaller models toward superior performance, verifying the framework's scalability and the transferability of its training-free logic.

\section{Related Work}
\subsection{LLM-based Optimization Modeling}
LLM-based optimization modeling reduces the expertise required for complex formulation via prompt-based strategies or fine-tuning methods.
Prompt-based frameworks utilize multi-agent workflows \cite{xiao2023chain, ahmaditeshnizioptimus, zhangcofft} or tree-search algorithms \cite{liuoptitree, astorgaautoformulation} to improve reasoning and code generation in general-purpose LLMs.
Conversely, fine-tuning methods like FOARL \cite{jianglarge}, ORLM \cite{huang2025orlm}, and SIRL \cite{chen2025solver} develop specialized models by training on domain-specific operations research datasets to internalize modeling patterns.

\subsection{Retrieval-Augmented Reasoning}
Despite the progress of reasoning LLMs \cite{openai2024learning, guo2025deepseek}, reliance on internal knowledge often leads to hallucinations that prompting alone cannot resolve \cite{huang2025survey, wei2022chain}. Retrieval-Augmented Reasoning addresses this by integrating external verification into the reasoning process \cite{barry2025graphrag}.
While frameworks like ReAct \cite{yao2022react} and Retrieval-Augmented Thoughts \cite{wang2024rat} dynamically revise traces with retrieved data, this paradigm is also effective for optimization.
For example, OptiTree \cite{liuoptitree} retrieves analogous subproblems to ground reasoning, ensuring complex modeling steps remain verifiable.

\begin{figure*}[t]
\centering
\includegraphics[width=0.99\textwidth]{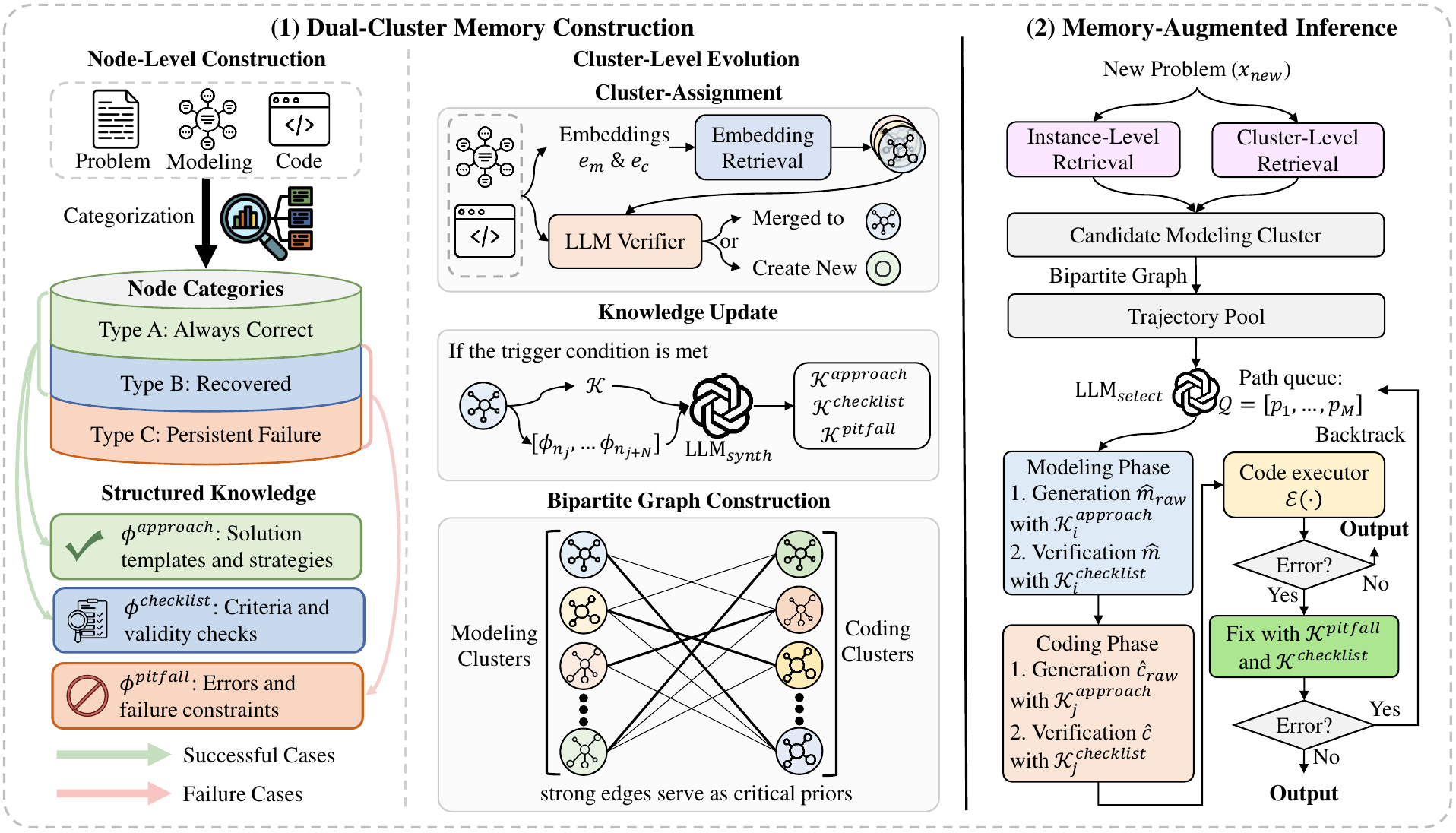}
\vspace{-3pt}
\caption{Overview of the Dual-Cluster Memory Agent (DCM-Agent).
This agent operates in two distinct phases: 
(1) \textbf{Dual-Cluster Memory Construction:} Historical solutions are stratified into three types to distill structured knowledge $\mathcal{K}$. These are organized into decoupled Modeling Clusters and Coding Clusters, bridged by a weighted bipartite graph $\mathcal{G}$. 
(2) \textbf{Memory-Augmented Inference:} For a new problem, the agent retrieves relevant cluster paths to guide the sequential generation of the mathematical formulation $\hat{m}$ and executable code $\hat{c}$.}
\label{fig:architecture}
\vspace{-10pt}
\end{figure*}


\section{Methodology}

\subsection{Overview}
We formalize optimization problem solving via LLMs as a composite structured process spanning the problem space $\mathcal{X}$, modeling space $\mathbb{M}$, and coding space $\mathbb{C}$. The solution $\hat{y}$ is derived via:
\begin{equation}
\hat{y} = \mathcal{E}(\underbrace{h_\psi(c \mid m)}_{\text{Coding}} \circ \underbrace{g_\phi(m \mid x)}_{\text{Modeling}})
\end{equation}
where $g_\phi$ generates the modeling logic $\hat{m}$, $h_\psi$ synthesizes the executable code $\hat{c}$, and $\mathcal{E}(\cdot)$ serves as the code executor.
The core challenge lies in the intrinsic one-to-many nature of this $x \to \hat{m} \to \hat{c}$ mapping: as a single problem admits diverse valid modeling logics and coding implementation (Figure \ref{fig:intro}), the model must possess robust judgment capabilities to navigate these possibilities.
\par
We propose the \textbf{Dual-Cluster Memory Agent (DCM-Agent)}, designed to manage this complexity by explicitly decoupling modeling ($g_\phi$) from coding ($h_\psi$), as shown in Figure \ref{fig:architecture}.
DCM-Agent operates in two phases: (1) \textit{Dual-Cluster Memory Construction}, which distills instance-level experience nodes into generalized structured knowledge, and (2) \textit{Memory-Augmented Inference}, which applies the knowledge to solve novel problems.
\par
To support this, DCM-Agent maintains a \textit{dynamic memory} $\mathcal{D}$ with a novel hierarchical structure: specific experience nodes are organized into decoupled Modeling Clusters and Coding Clusters.
While individual nodes store trajectory details, each cluster maintains a high-level generalized knowledge ($\mathcal{K}$) synthesized from its constituent nodes.
At the same time, DCM-Agent constructs a specialized bipartite graph $\mathcal{G}$ to bridge these clusters, modeling the compatibility between abstract modeling logics and concrete coding strategies.

\subsection{Dual-Cluster Memory Construction}
This phase transforms raw problem-solution trajectories into a structured, decoupled memory system. 
The process follows a rigorous bottom-up lifecycle, progressing from the discrete analysis of individual instances to collective knowledge synthesis.

\subsubsection{Node-Level Construction}
\label{sec:extraction}
We first curate a dataset of distinct samples, disjoint from the evaluation benchmarks.
To ensure the quality of extracted knowledge, we classify the solutions of these samples into three categories:
\begin{itemize}[leftmargin=*]
\item \textbf{Type A (Always Correct):}
Samples are consistently solved correctly across multiple attempts. 
These represent canonical problem-solving patterns and serve as standard references.

\item \textbf{Type B (Recovered):}
The samples that initially failed but yielded correct solutions upon re-attempting. 
These instances capture the boundary between failure and success, providing critical information on specific failure modes and their corresponding recovery reasoning.

\item \textbf{Type C (Persistent Failure):}
The solutions persist in failure despite exhaustive attempts (e.g., exceeding 3 rounds). 
These encode fundamental mismatches between problems and approaches.
\end{itemize}
This stratification facilitates differential knowledge extraction: the successful instances (Types A and B) provide positive references and caveats, while failures (Types B and C) reveal critical pitfalls.
\par
To transform these raw samples into actionable memory, we first decompose each problem-solution pair into distinct modeling logic and coding implementation components.
For each component, we generate embeddings ($\mathbf{e}_m, \mathbf{e}_c$) to enable semantic retrieval, and simultaneously extract a tuple of instance-specific knowledge, denoted as $\Phi_n = \langle \phi_n^{\text{approach}}, \phi_n^{\text{checklist}}, \phi_n^{\text{pitfall}} \rangle$, as summarized in Table~\ref{tab:extraction}.
Here, $\Phi_n$ represents local insights specific to node $n$.
Specifically, we synthesize canonical approaches ($\phi_n^{\text{approach}}$) and verification checklists ($\phi_n^{\text{checklist}}$) from successful nodes (Types A and B), while deriving explicit pitfall warnings ($\phi_n^{\text{pitfall}}$) from failure trajectories (Types B and C).

\begin{table}[t]
\centering
\resizebox{\linewidth}{!}{%
\begin{tabular}{@{}lll@{}}
\toprule
\textbf{Guidance Tier} & \textbf{Source Type} & \textbf{Content Definition} \\
\midrule
$\phi^{\text{approach}}$ & Type A + Type B & Solution templates \\
(How to solve) & (Success) & and logical steps \\
\midrule
$\phi^{\text{checklist}}$ & Type A + Type B & Validity criteria and \\
(What to verify) & (Success) & boundary checks \\
\midrule
$\phi^{\text{pitfall}}$ & Type B + Type C & Common errors and \\
(What to avoid) & (Failure) & constraint violations \\
\bottomrule
\end{tabular}
}
\vspace{-3pt}
\caption{Mapping from sample types to guidance tiers.}
\label{tab:extraction}
\vspace{-10pt}
\end{table}

\subsubsection{Cluster-Level Evolution}
\label{sec:evolution}
Once individual nodes are processed, DCM-Agent organizes them to form generalized knowledge.
 This involves clustering nodes, synthesizing knowledge, and establishing graph connectivity.

\textbf{Cluster Assignment.}
Each experience node is integrated into the memory via a rigorous assignment process.
First, embedding-based retrieval identifies top-$k$ candidate clusters by comparing the node's embedding ($\mathbf{e}_m$ or $\mathbf{e}_c$) with cluster centroids $\boldsymbol{\mu}$.
Second, an LLM-based verifier checks semantic consistency to either merge the node into a matched candidate or initialize a new cluster.

\begin{figure}[t]
\centering
\includegraphics[width=0.47\textwidth]{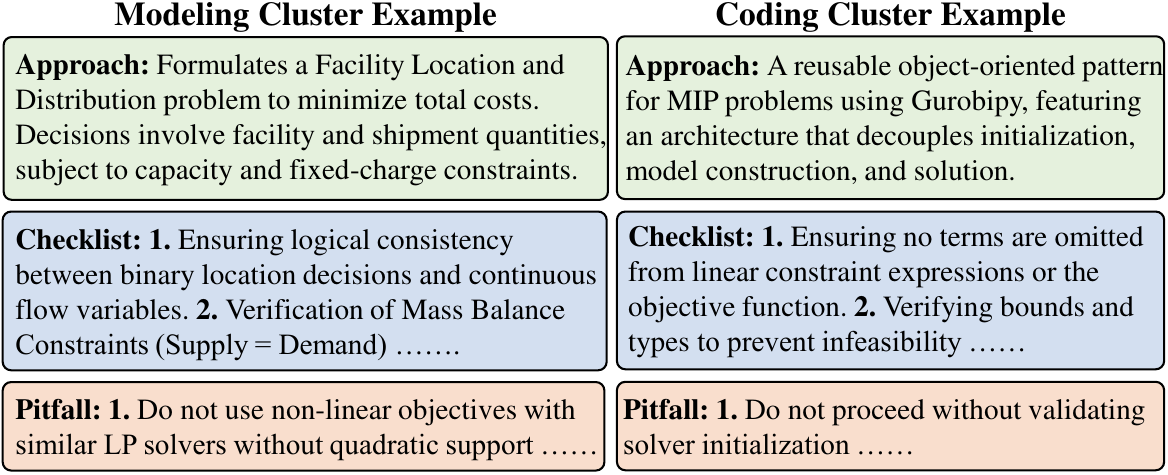}
\vspace{-3pt}
\caption{Two examples in our Dual-Cluster Memory.}
\label{fig:memory_examle}
\vspace{-12pt}
\end{figure}

\textbf{Knowledge Update.}
To resolve the ambiguity between the specific examples and general patterns, we employ an incremental update mechanism.
Each cluster maintains a generalized knowledge $\mathcal{K}=\langle \mathcal{K}^{\text{approach}}, \mathcal{K}^{\text{checklist}}, \mathcal{K}^{\text{pitfall}} \rangle$, which serves as the consolidated schema for that cluster (distinct from the raw $\Phi_n$ of individual nodes).
When a cluster accumulates a threshold of new nodes (such as $N=5$), we trigger a knowledge update step:
\begin{equation}
\mathcal{K}^{(t+1)} = \text{LLM}_{\text{synth}}\left(\mathcal{K}^{(t)} \ \cup \ \bigcup_{j=1}^{N} \Phi_{n_j} \right)
\end{equation}
Here, $\text{LLM}_{\text{synth}}$ is used to abstract generalized patterns from the new batch of instance knowledge $\Phi_{n_j}$ and merge them into the generalized knowledge $\mathcal{K}^{(t)}$.
This ensures that $\mathcal{K}$ evolves to capture robust, non-redundant insights while retaining specific pitfall warnings, and is not overly influenced by extreme samples, as shown in Figure \ref{fig:memory_examle}.

\textbf{Bipartite Graph Construction.}
At the same time, we introduce a bipartite graph $\mathcal{G}$ to model the associations between the these decoupled clusters.
Since each experience node $n$ naturally maps to a pair of clusters $(C^M_i, C^C_j)$, these linkages aggregate into a global structure.
We formalize this as a bipartite graph $\mathcal{G} = (V_{M}, V_{C}, E)$, where the edge weight $w_{ij}$ quantifies the co-occurrence frequency of modeling logic $C^M_i$ and coding strategy $C^C_j$.
The strong edges represent proven pathways, providing critical priors for subsequent usage.

\subsection{Memory-Augmented Inference}
\subsubsection{Dual-Retrieval}
For a new problem $x_{\text{new}}$, DCM-Agent leverages the memory to efficiently navigate the solution space by retrieving relevant historical experiences.
We first encode the problem into the modeling logic embedding $\mathbf{e}_{\text{new}}$ and employ two complementary retrieval mechanisms to balance the problem relevance with general algorithmic applicability:
\par
\textbf{Instance-Level Retrieval} captures the granular problem similarity by retrieving specific nodes $\mathcal{H}$ closest to $\mathbf{e}_{\text{new}}$, thereby identifying relevant experience nodes that share detailed semantic features:
\begin{equation}
\mathcal{H} = \arg\max_{K} \{\text{sim}(\mathbf{e}_{\text{new}}, \mathbf{e}_i) \mid x_i \in \mathcal{D}\}.
\end{equation}
\par
\textbf{Cluster-Level Retrieval} targets abstract patterns by comparing $\mathbf{e}_{\text{new}}$ directly with cluster centroids $\boldsymbol{\mu}_k^M$, which ensures capturing the modeling logic beyond surface-level textual matches:
\begin{equation}
\mathcal{S}_{\text{cluster}} = \arg\max_{K} \{\text{sim}(\mathbf{e}_{\text{new}}, \boldsymbol{\mu}_k^M)\}
\end{equation}
\par
These two sources are united to yield a robust final set of candidate modeling clusters, integrating both specific exemplars and general categories:
\begin{equation}
\mathcal{R} = \{C^M(x_i) \mid x_i \in \mathcal{H}\} \cup \mathcal{S}_{\text{cluster}}.
\end{equation}
\par
To effectively bridge modeling logic with coding implementation, we query the graph $\mathcal{G}$ to exploit learned associations. 
For each identified $C^M_i \in \mathcal{R}$, we retrieve the top-$K$ coding neighbors $\mathcal{N}_k$ with the highest edge weights to form a diverse trajectory pool $\mathcal{P}$.
Finally, an LLM selector serves as a verifier to rank these combinations based on their logical alignment with $x_{\text{new}}$, returning a prioritized queue $\mathcal{Q}$ of the most promising $M$ solution paths:
\begin{equation}
\mathcal{Q} = \mathop{\text{Top-}M}_{(p) \in \mathcal{P}} \left( \text{LLM}_{\text{select}}(\mathcal{P}, x_{\text{new}}) \right)
\end{equation}

\subsubsection{Solving via Generalized Knowledge}
DCM-Agent processes the prioritized queue $\mathcal{Q}$ using a Generate-Verify-Repair-Backtrack pipeline. 
Crucially, all the steps are conditioned on the $\mathcal{K}$ of the selected clusters, rather than node knowledge $\phi$, as shown in  Figure \ref{fig:architecture}.
For a path $p_t = (C^M_i, C^C_j)$:
\par
\textbf{1. Generation \& Verification.}
The LLM generates the modeling logic $\hat{m}_{\text{raw}}$ using the cluster's canonical approach $\mathcal{K}_i^{\text{approach}}$.
Immediately, it verifies $\hat{m}_{\text{raw}}$ against the cluster's checklist $\mathcal{K}_i^{\text{checklist}}$:
\begin{align}
\hat{m}_{\text{raw}} &= \text{LLM}_{\text{gen}}(x_{\text{new}} \mid \mathcal{K}_i^{\text{approach}}) \\
\hat{m} &= \text{LLM}_{\text{verify}}(\hat{m}_{\text{raw}} \mid \mathcal{K}_i^{\text{checklist}})
\end{align}
Once the $\hat{m}$ is established, DCM-Agent transitions to code generation.
The executable code $\hat{c}$ is generated using the coding cluster's templates $\mathcal{K}_j^{\text{approach}}$ and rigorously checked against guidelines $\mathcal{K}_j^{\text{checklist}}$, ensuring the robustness of the code structure

\textbf{2. Repair \& Backtracking.}
If the execution $\mathcal{E}(\hat{c})$ fails, resulting in the {runtime} error $e$, DCM-Agent initiates a knowledge-guided repair mechanism.
Instead of blind debugging, it systematically analyzes $e$ in the context of the cluster's specific pitfall warnings $\mathcal{K}_j^{\text{pitfall}}$, which contain common error patterns associated with this algorithm type:
\begin{equation}
\hat{c}_{\text{fixed}} = \text{LLM}_{\text{fix}}(\hat{c}, e \mid \mathcal{K}_j^{\text{pitfall}}, \mathcal{K}_j^{\text{checklist}})
\end{equation}
If repair attempts fail to yield a solution within a limit, the agent triggers a backtracking protocol. 
It discards the path and activates the next $p_{t+1}$ from $\mathcal{Q}$, preventing the system from getting stuck in local optima and ensuring robust problem-solving.

\begin{table*}[t]
\centering
\resizebox{\textwidth}{!}{%
\begin{tabular}{lcccccccc}
\toprule
\multirow{2}{*}{\textbf{Method}} & \multicolumn{7}{c}{\textbf{Datasets}} & \multirow{2}{*}{\textbf{Avg.}} \\
\cmidrule(lr){2-8}
 & \textbf{NL4Opt} & \textbf{ComplexLP} & \textbf{NLP4LP} &\textbf{OptiBench} & \textbf{OptMATH} & \textbf{IndOR} & \textbf{ComplexOR} & \\
 \textbf{Size} & 230 & 211 & 242 & 605 & 166 & 100 & 18 & -\\
 \midrule
  \rowcolor{gray!20} \multicolumn{9}{c}{\textbf{Qwen3-8B}} \\
Baseline & 41.74 & 16.11 & 35.12 & 30.58 & 10.24 & 19.00 & 22.22 & 27.99 \\
OptiMUS & 53.48 & 23.22 & 43.39 & 44.13 & 15.06 & 23.00 & 33.33 & 38.04 \\
AF-MCTS & 47.39 & 19.43 & 39.26 & 36.53 & 12.65 & 21.00 & 33.33 & 32.70 \\
OptiTree & 55.22 & 25.59 & 46.28 & 45.12 & 16.26 & 25.00 & 38.89 & 39.76 \\
\textbf{DCM-Agent} & \textbf{64.35} & \textbf{32.23} & \textbf{62.40} & \textbf{55.37} & \textbf{21.69} & \textbf{30.00} & \textbf{50.00} & \textbf{49.43} \\
 \midrule
 \rowcolor{gray!20} \multicolumn{9}{c}{\textbf{Qwen3-30B}} \\
Baseline & 55.22 & 28.44 & 52.07 & 43.64 & 16.87 & 25.00 & 38.89 & 40.52 \\
OptiMUS & 63.48 & 33.64 & 63.64 & 56.20 & 24.70 & 29.00 & 50.00 & 50.25 \\
AF-MCTS & 65.65 & 34.60 & 68.18 & 57.68 & 25.90 & 31.00 & 50.00 & 52.22 \\
OptiTree & 70.88 & 36.49 & 70.25 & 59.50 & 27.71 & 30.00 & 55.56 & 54.45 \\
\textbf{DCM-Agent} & \textbf{77.39} & \textbf{41.23} & \textbf{76.03} & \textbf{64.30} & \textbf{32.53} & \textbf{34.00} & \textbf{61.11} & \textbf{59.61} \\
 \midrule
 \rowcolor{gray!20} \multicolumn{9}{c}{\textbf{Qwen3-235B}} \\
Baseline & 78.13 & 40.76 & 74.38 & 60.83 & 31.33 & 32.00 & 55.56 & 57.74 \\
OptiMUS & 83.48 & 44.55 & 77.27 & 64.30 & 37.95 & 34.00 & 66.67 & 61.77 \\
AF-MCTS & 87.39 & 46.92 & 78.93 & 68.26 & 40.36 & 37.00 & 61.11 & 64.82 \\
OptiTree & 89.56 & 48.82 & 80.16 & 70.74 & 41.57 & 36.00 & 66.67 & 66.66 \\
\textbf{DCM-Agent} & \textbf{93.48} & \textbf{54.76} & \textbf{84.71} & \textbf{75.21} & \textbf{46.39} & \textbf{40.00} & \textbf{72.22} & \textbf{71.28} \\
 \midrule
 \rowcolor{gray!20} \multicolumn{9}{c}{\textbf{Deepseek-V3.2}} \\
Baseline & 82.61 & 38.39 & 71.73 & 58.68 & 36.75 & 34.00 & 61.11 & 57.61 \\
OptiMUS & 86.09 & 43.60 & 75.21 & 62.15 & 39.76 & 36.00 & 66.67 & 61.20 \\
AF-MCTS  & 89.13 & 46.92 & 78.10 & 67.27 & 43.98 & 39.00 & 66.67 & 65.14 \\
OptiTree & 90.87 & 47.87 & 79.34 & 70.08 & 47.59 & 38.00 & 72.22 & 67.18 \\
\textbf{DCM-Agent} & \textbf{95.22} & \textbf{53.55} & \textbf{83.06} & \textbf{74.05} & \textbf{52.73} & \textbf{41.00} & \textbf{77.77} & \textbf{71.47} \\
 \midrule
 \rowcolor{gray!20} \multicolumn{9}{c}{\textbf{GPT5.1}} \\
Baseline & 87.39 & 55.45 & 84.30 & 68.26 & 43.38 & 44.00 & 66.67 & 67.62 \\
OptiMUS & 90.43 & 58.77 & 86.78 & 71.07 & 45.78 & 46.00 & 72.22 & 70.42 \\
AF-MCTS & 94.47 & 60.66 & 88.84 & 74.71 & 51.20 & 51.00 & 77.77 & 73.93 \\
OptiTree & 95.65 & 62.56 & 90.08 & 76.86 & 53.61 & 49.00 & 77.77 & 75.51 \\
\textbf{DCM-Agent} & \textbf{97.83} & \textbf{65.40} & \textbf{93.39} & \textbf{80.16} & \textbf{55.42} & \textbf{53.00} & \textbf{83.33} & \textbf{78.50} \\
\bottomrule
\end{tabular}
}
\vspace{-3pt}
\caption{Solving accuracy (\%) comparison across different datasets. The best results are highlighted in \textbf{bold}.}
\label{tab:main_results}
\vspace{-7pt}
\end{table*}

\begin{table}[t]
\centering
\resizebox{\linewidth}{!}{%
\begin{tabular}{lccc}
\toprule
\textbf{Method} & \textbf{NLP4LP} & \textbf{OptiBench} & \textbf{OptMATH} \\
\midrule
Baseline & 8.3s  & 13.7s & 21.7s  \\
OptiMUS  & 26.5s & 46.6s & 86.3s  \\
AF-MCTS  & 85.3s & 110.8s & 205.7s \\
OptiTree & 17.7s & 33.5s & 61.6s  \\
DCM-Agent  & 22.1s & 41.3s & 73.4s \\
\bottomrule
\end{tabular}
}
\caption{Time cost statistics (in seconds) of the Qwen3-235B model across selected benchmarks.}
\label{tab:time_cost_235b_transposed}
\vspace{-10pt}
\end{table}

\section{Experiment}
\subsection{Experiment Setups}
\textbf{Datasets.}
To comprehensively evaluate our DCM-Agent framework across a spectrum of complexities, we utilize a diverse suite of seven optimization benchmarks. 
We employ {NL4Opt} \cite{ramamonjison2023nl4opt} and {NLP4LP} \cite{ahmaditeshnizioptimus} as standard baselines for linear and mixed-integer programming. 
To ensure rigorous testing on more demanding tasks, we incorporate {OptiBench} \cite{yangoptibench},  {OptMATH} \cite{luoptmath}, and the {ComplexLP} subset of {MAMO} \cite{huang-etal-2025-llms}. 
Finally, we assess real-world applicability using IndustryOR \cite{huang2025orlm} and ComplexOR \cite{xiao2023chain} datasets.
The memory is constructed using $500$ problems randomly sampled from the mathematical programming and combinatorial optimization domains within OptiVerse benchmark \cite{zhang2026optiverse}, ensuring no intersection with the evaluation benchmarks.
\par
\textbf{Baselines.}
We systematically compare DCM-Agent with diverse methods on different LLMs of varying sizes, ranging from generic standard LLMs to advanced specialized optimization frameworks.
We first establish a fundamental baseline using LLMs of varying sizes, including Qwen3 series (8B, 30B, 235B) \cite{yang2025qwen3}, DeepSeek-V3.2 \cite{liu2025deepseek}, and GPT-5.1 \cite{openai2025gpt51}.
Subsequently, we compare against representative specialized methods:
(1) OptiMUS \cite{ahmaditeshnizioptimus}, a multi-agent workflow that enhances reliability through structured input processing;
(2) AF-MCTS \cite{astorgaautoformulation}, which employs Monte Carlo Tree Search to sequentially identify variables, constraints, and objectives; and
(3) OptiTree \cite{liuoptitree}, which tackles high-complexity tasks by adaptively decomposing problems into manageable sub-problems.
\par
\textbf{Evaluation.}
Consistent with prior research, we adopt strict \textit{end-to-end solving accuracy} as our primary evaluation metric.
Our protocol evaluates the complete resolution pipeline: given a problem, the model generates the code $c$ to produce execution output $o = \text{Python}(c)$, where the allowed libraries are \textit{Gurobi}, \textit{PuLP}, \textit{OR-Tools}, \textit{SciPy}, and \textit{NetworkX}.
A problem is considered solved if the extracted numerical answers for both the requirement and the objective function match the ground truth.

\begin{table}[t]
\centering
\resizebox{\columnwidth}{!}{%
\begin{tabular}{lccccc}
\toprule
\textbf{Ratio} & 0\% & {10\%} & {40\% } & {70\%}  & {100\%}\\
\midrule
NLP4LP & 74.38 & 78.51 & 81.40  & 82.64  & 84.71\\
OptiBench  & 60.83 & 66.45 & 71.57 & 74.05 &  75.21\\
OptMATH  & 31.33 & 36.75 & 40.36  & 44.58 &  46.39\\
\bottomrule
\end{tabular}
}
\vspace{-3pt}
\caption{Performance comparison under different memory budgets (10\%, 40\%, 70\%, and 100\%) with Qwen3-235B across selected datasets. The results demonstrate the impact of memory constraints on solving accuracy.}
\label{tab:memory_ablation}
\vspace{-12pt}
\end{table}

\begin{table}[t]
\centering
\resizebox{\columnwidth}{!}{%
\begin{tabular}{lccc}
\toprule
Memory  & {NLP4LP}  &{OptiBench} & {OptMATH} \\
\midrule
 \rowcolor{gray!20} \multicolumn{4}{c}{\textbf{Qwen3-8B}} \\
Baseline & {35.12}  & {30.58} & {10.24}  \\
Qwen3-8B & {62.40} & {55.37} & {21.69}  \\
Qwen3-30B & {67.36}  & {59.34} & {23.49}  \\
Qwen3-235B & \textbf{69.42} & \textbf{60.17} & \textbf{25.30}  \\
DeepSeek-V3.2 & {66.53}  & {58.68} & {24.10}  \\
GPT-5.1 & {65.28} & {57.52} & {22.89}  \\
 \midrule
 \rowcolor{gray!20} \multicolumn{4}{c}{\textbf{Qwen3-235B}} \\
Baseline & {74.38}  & {60.83} & {31.33}  \\
Qwen3-8B & 78.51 & 69.59 & 40.96  \\
Qwen3-30B & 82.64 & 73.22 & 43.98  \\
Qwen3-235B & {84.71}  & {75.21} & {46.39}  \\
DeepSeek-V3.2 & {85.54}  & {76.03} & {47.59}  \\
GPT-5.1 & \textbf{86.36}  & \textbf{77.36} & \textbf{48.19}  \\
\bottomrule
\end{tabular}
}
\vspace{-3pt}
\caption{Cross-model knowledge transfer: Impact of memory construction model on performance (\%).}
\label{tab:result_1}
\vspace{-1pt}
\end{table}

\begin{figure}[t]
\centering
\includegraphics[width=0.985\linewidth]{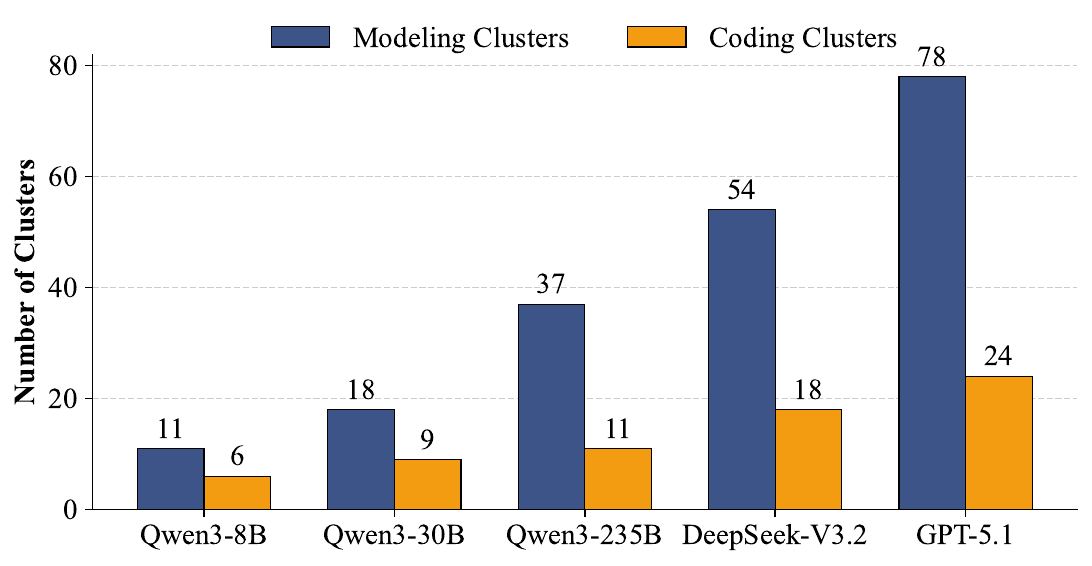}
\vspace{-3pt}
\caption{Statistical distribution of the number of two clusters obtained from LLMs of different sizes.}
\label{fig:cluster}
\vspace{-10pt}
\end{figure}

\subsection{Main Results}
By carefully analyzing the accuracy results in Table~\ref{tab:main_results} alongside the detailed time cost statistics in Table~\ref{tab:time_cost_235b_transposed}, we derive the following key insights:

\textbf{Superiority of Our DCM-Agent across Model Scales.} 
DCM-Agent consistently achieves the highest accuracy across all evaluated model sizes, ranging from the 8B parameter scale to GPT5.1. 
This uniform success demonstrates the framework's robustness and its capability to serve as a universal performance enhancer for optimization tasks regardless of the underlying backbone architecture.

\textbf{Lower Capability Requirements for Memory Construction.} 
The performance gains of DCM-Agent are most pronounced on smaller LLMs because its structured memory construction process is relatively less demanding on the model's inherent reasoning power. 
By offloading constraint management to an external memory module, DCM-Agent allows smaller models to overcome their parameter limitations and thereby achieve competitive results typically reserved for much larger models.

\textbf{Balanced Efficiency and Computational Overhead.} 
As shown in Table~\ref{tab:time_cost_235b_transposed}, DCM-Agent significantly reduces the cost time compared to heavy search-based methods like AF-MCTS while maintaining superior accuracy. 
This suggests that DCM-Agent’s memory-driven reasoning is more computationally purposeful than exhaustive tree exploration, providing an optimal overall trade-off between solving performance and time efficiency.

\begin{figure}[t]
\centering
\includegraphics[width=0.47\textwidth]{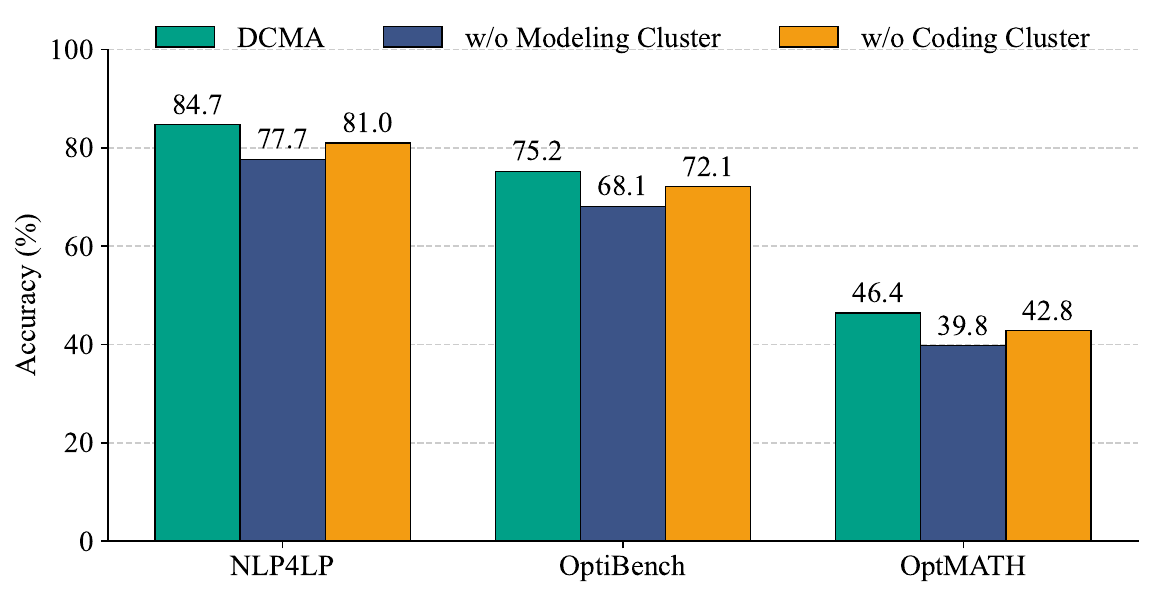}
\vspace{-3pt}
\caption{Ablation study on NLP4LP, OptiBench, and OptMATH datasets using Qwen3-235B.}
\label{fig:abalation}
\vspace{-3pt}
\end{figure}


\begin{table}[t]
\centering
\resizebox{0.9\linewidth}{!}{
\begin{tabular}{lccc}
\toprule
{Setting} & {NLP4LP} & {OptiBench} & {OptMATH} \\
\midrule
\multicolumn{4}{l}{\textit{Retrieval Top-$K$ in construction and inference}} \\
$K=1$ & 79.34 & 70.91 & 40.36 \\
$K=3$ & {84.71} & {75.21} & {46.39} \\
$K=5$ & 83.06 & 74.05 & 43.98 \\
\midrule
\multicolumn{4}{l}{\textit{Memory Update Threshold ($N$) in construction}} \\
$N=1$ & 84.30 & 74.21 & 44.58 \\
$N=5$ & {84.71} & {75.21} & {46.39} \\
$N=10$ & 82.64 & 72.73 &  42.77\\
\midrule
\multicolumn{4}{l}{\textit{Planning Candidates ($M$) in inference}} \\
$M=1$ & 82.23 & 72.56 & 42.17 \\
$M=3$ & {84.71} & {75.21} & {46.39} \\
$M=5$ & 85.54 & 75.87 & 46.99 \\
\bottomrule
\end{tabular}
}
\vspace{-3pt}
\caption{Parameter sensitivity analysis on key hyperparameters ($K$, $M$, $N$) across three benchmarks.}
\label{tab:params}
\vspace{-10pt}
\end{table}


\begin{figure*}[t]
\centering
\includegraphics[width=\textwidth]{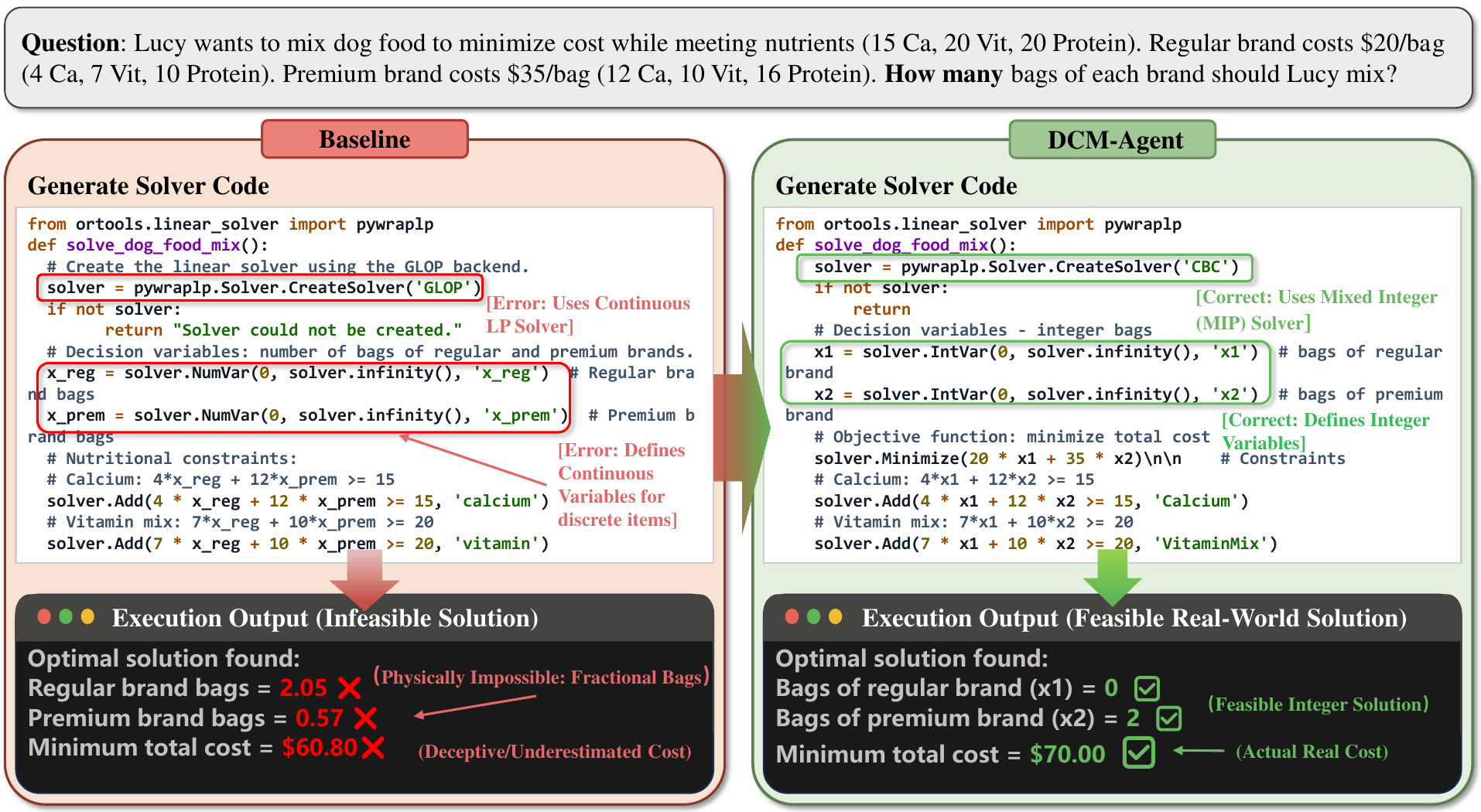}
\vspace{-11pt}
\caption{Comparison between the baseline and DCM-Agent on a discrete optimization task. DCM-Agent correctly identifies integrality constraints, whereas the baseline produces a physically infeasible fractional solution.}
\label{fig:case_study}
\vspace{-10pt}
\end{figure*}

\subsection{Dual-Cluster Memory Analysis}
\label{subsec:memory_analysis}
\textbf{Cluster Statistics across LLMs.}
We investigate the number of Modeling Clusters (MC) and Constraint Clusters (CC) generated by different LLMs during the memory construction phase. As illustrated in Figure \ref{fig:cluster}, stronger models (e.g., GPT-5.1) tend to generate a significantly higher number of Modeling Clusters compared to smaller models (e.g., Qwen3-8B). 
This suggests that advanced LLMs possess a more granular understanding of abstract problem structures, allowing them to disentangle subtle differences in mathematical formulations that smaller models might otherwise conflate.
\par
\textbf{Impact of Memory Nodes.}
We assess the system's robustness by varying the available memory budget from 0\% to 100\% using Qwen3-235B as the backbone. 
As shown in Table \ref{tab:memory_ablation}, performance improves progressively as the number of memory nodes increases. 
This confirms that the breadth of historical experiences directly correlates with the system's ability to generalize to novel problems.
\par
\textbf{Cross-Model Memory Transferability.}
A pivotal advantage of DCM-Agent lies in the architectural decoupling of memory construction from inference, enabling flexible cross-model synergy.
As shown in Table \ref{tab:result_1}, we observe a compelling ``knowledge inheritance'' effect: structural priors generated by superior models significantly elevate the performance of smaller counterparts.
However, performance eventually declines as memory-generating LLMs become exceptionally powerful; we hypothesize that the resulting high-density cluster information exceeds the processing capacity of smaller models.
This confirms that while weaker models can inherit superior structural reasoning from stronger ones, the complexity remains within the target model's processing threshold.



\subsection{Ablation Studies}
To assess the impact of DCM-Agent's components, we conduct ablation studies using Qwen3-235B by selectively removing the Modeling and Coding Clusters.
As illustrated in Figure \ref{fig:abalation}, the complete framework consistently yields the highest accuracy, validating the synergy of the dual-cluster mechanism.
We observe that the removal of Modeling Clusters results in a sharper performance drop compared to removing Coding Clusters (e.g., the blue bars are consistently lower than the orange ones).
This result confirms that while executable code guidance is beneficial, decoupling and retrieving precise mathematical logic is the decisive factor for solving complex optimization problems.

\subsection{Parameter Analysis}
We systematically evaluate DCM-Agent's sensitivity to retrieval size ($K$), update threshold ($N$), and planning candidates ($M$), summarized in Table \ref{tab:params}.
Both $K$ and $N$ exhibit a distinct bell-shaped trend, peaking at $K=3$ and $N=5$.
Lower values suffer from insufficient context or unstable generalization (overfitting to isolated samples), while higher values are hampered by excessive input context length or delayed knowledge consolidation.
In contrast, increasing $M$ yields consistent monotonic gains by broadening the search space.
However, given the marginal improvement at $M=5$ relative to the incurred computational overhead, we adopt $M=3$ to ensure the optimal cost-effectiveness.



\subsection{Case Study}
As shown in Figure \ref{fig:case_study}, DCM-Agent identifies implicit integrality constraints (e.g., discrete bag quantities) and selects an appropriate Mixed-Integer Programming (MIP) solver. 
Conversely, the baseline overlooks these constraints due to semantic misalignment, treating the problem as a continuous linear task. 
While the baseline reports a nominally lower cost (\$60.80 vs. \$70.00), its reliance on fractional quantities results in an infeasible solution. 
This ``deceptive optimality'' underscores that numerical performance is secondary to the foundational correctness of the modeling logic, such as the proper distinction between \texttt{IntVar} and \texttt{NumVar}.

\section{Conclusion}
We presented the novel Dual-Cluster Memory Agent (DCM-Agent), which addresses structural ambiguity in optimization by decoupling abstract modeling logic from concrete code implementation.
By leveraging Dual-Cluster Organization and Evolutionary Experience Stratification, DCM-Agent provides precise algorithm-specific guidance through stratified historical insights.
Empirical results across seven benchmarks confirm that DCM-Agent achieves consistent state-of-the-art accuracy across various model scales while not introducing too much computational overhead.
This framework effectively bridges the critical gap between mathematical formulation and executable solver code, significantly improving overall solution robustness.
DCM-Agent thus offers a scalable and efficient approach to navigating the highly complex decision space of automated optimization problem solving.

\section{Acknowledgements}
This work was supported by Fundamental and Interdisciplinary Disciplines Breakthrough Plan of the Ministry of Education of China (JYB2025XDXM116), National Natural Science Foundation of China (No. 62137002, 62293550, 62293553, 62293554, 62437002, 62477036, 62477037, 62192781),  the Shaanxi Provincial Social Science Foundation Project (No. 2024P041), the Youth Innovation Team of Shaanxi Universities "Multi-modal Data Mining and Fusion", and Xi'an Jiaotong University City College Research Project (No. 2024Y01).

\section{Limitation} 
Despite the superior performance of the Dual-Cluster Memory Agent (DCM-Agent) across various benchmarks and model scales, we identify one primary limitation concerning the framework's initialization process. 
A notable bottleneck lies in the initialization latency during the memory construction phase. Specifically, distilling structured knowledge requires collecting and classifying historical trajectories into distinct categories and progressively building the bipartite graph. 
This process incurs a one-time computational overhead that is non-negligible compared to zero-shot prompting.
However, it is crucial to emphasize that this represents a "sunk cost": once the Dual-Cluster Memory is fully constructed and stabilized, the framework operates in a plug-and-play manner. The subsequent Memory-Augmented Inference phase is highly efficient, relying on fast embedding retrieval and path planning rather than exhaustive re-generation. This design effectively amortizes the initial construction cost over long-term usage.
In future work, we aim to explore online learning mechanisms that enable the memory to evolve dynamically through user interactions, eliminating the need for full reconstruction cycles.

\section{Ethical Statement}
In developing DCM-Agent, we have rigorously considered the ethical implications of our research, particularly concerning data integrity, privacy, and computational sustainability.
Our experiments utilize established, publicly available optimization benchmarks (including NL4Opt, OptiBench, and NLP4LP) and our own collection of 500 questions, which consist solely of mathematical optimization problems and their solutions. We have verified that these datasets do not contain personally identifying information, sensitive data, or offensive content. The problems are purely technical in nature, involving mathematical formulations and programming tasks without any reference to individual identities or potentially harmful content.
We strictly adhere to the licensing agreements of these datasets and the large language models employed (e.g., Qwen, DeepSeek, GPT). By demonstrating that smaller models (e.g., 8B parameters) can achieve state-of-the-art performance when augmented with our memory system—often surpassing unaugmented larger models—DCM-Agent offers a pathway to reduce the substantial energy consumption typically associated with running massive foundation models for complex reasoning tasks. This contribution aligns with the broader goal of developing more sustainable and accessible AI systems.

\bibliography{custom}


\appendix
\section{Details of Ai Assistants In Research Or Writing}
We used Claude-4.5-Sonnet and Gemini-3.0-Pro to help us write code and polish the paper.

\begin{algorithm*}[!ht]
\caption{Dual-Cluster Memory Construction}
\label{alg:memory-construction}
\begin{algorithmic}[1]
\Require Historical problem-solution pairs $\mathcal{D}_{raw}$
\Ensure Modeling clusters $\mathcal{C}^M$, Coding clusters $\mathcal{C}^C$, Bipartite graph $\mathcal{G}$

\State \textbf{// Phase 1: Node-Level Construction}
\For{each problem-solution pair $(x_i, y_i) \in \mathcal{D}_{raw}$}
    \State Classify into Type A/B/C based on solving attempts
    \State Decompose into modeling logic $m_i$ and coding implementation $c_i$
    \State Generate embeddings: $e_i^m \leftarrow \text{Embed}(m_i)$, $e_i^c \leftarrow \text{Embed}(c_i)$
    \State Extract instance-level knowledge:
    \State \quad $\phi_i^{approach} \leftarrow \text{Extract}_{\text{approach}}(\text{Type A/B})$
    \State \quad $\phi_i^{checklist} \leftarrow \text{Extract}_{\text{checklist}}(\text{Type A/B})$
    \State \quad $\phi_i^{pitfall} \leftarrow \text{Extract}_{\text{pitfall}}(\text{Type B/C})$
    \State $\Phi_i \leftarrow \langle \phi_i^{approach}, \phi_i^{checklist}, \phi_i^{pitfall} \rangle$
\EndFor

\State \textbf{// Phase 2: Cluster-Level Evolution}
\For{each node $n_i$ with $(e_i^m, e_i^c, \Phi_i)$}
    \State \textbf{// Modeling Cluster Assignment}
    \State $\mathcal{C}_{cand}^M \leftarrow \text{TopK-Retrieve}(e_i^m, \{\mu_k^M\})$ \Comment{k candidate clusters}
    \If{$\text{LLM}_{\text{verify}}(m_i, \mathcal{C}_{cand}^M)$ returns match at cluster $\mathcal{C}_j^M$}
        \State Add $n_i$ to $\mathcal{C}_j^M$
    \Else
        \State Create new cluster $\mathcal{C}_{new}^M$ with centroid $\mu_{new}^M = e_i^m$
    \EndIf
    
    \State \textbf{// Coding Cluster Assignment (similar process)}
    \State Assign $c_i$ to coding cluster $\mathcal{C}_k^C$
    
    \State \textbf{// Knowledge Update}
    \If{$|\mathcal{C}_j^M|_{new} \geq N$} \Comment{Threshold reached}
        \State $\mathcal{K}_j^M \leftarrow \text{LLM}_{\text{synth}}\left(\mathcal{K}_j^{M(t)} \cup \bigcup_{n \in \text{new}} \Phi_n\right)$
    \EndIf
    
    \State \textbf{// Bipartite Graph Update}
    \State Update edge weight: $w_{jk} \leftarrow w_{jk} + 1$ for $(\mathcal{C}_j^M, \mathcal{C}_k^C)$
\EndFor

\Return $\mathcal{C}^M, \mathcal{C}^C, \mathcal{G} = (\mathcal{V}^M, \mathcal{V}^C, \mathcal{E})$
\end{algorithmic}
\end{algorithm*}

\begin{algorithm*}[!ht]
\caption{Memory-Augmented Inference}
\label{alg:memory-inference}
\begin{algorithmic}[1]
\Require New problem $x_{new}$, Memory $\mathcal{D}$, Bipartite graph $\mathcal{G}$
\Ensure Solution $\hat{y}$ (modeling $\hat{m}$ and code $\hat{c}$)

\State \textbf{// Phase 1: Dual-Retrieval}
\State Encode problem: $e_{new} \leftarrow \text{Embed}(x_{new})$

\State $\mathcal{H} \leftarrow \arg\max_K \{\text{sim}(e_{new}, e_i) | x_i \in \mathcal{D}\}$ \Comment{Instance-Level Retrieval}

\State $\mathcal{S}_{cluster} \leftarrow \arg\max_K \{\text{sim}(e_{new}, \mu_k^M)\}$ \Comment{Cluster-Level Retrieval}

\State $\mathcal{R} \leftarrow \{\mathcal{C}^M(x_i) | x_i \in \mathcal{H}\} \cup \mathcal{S}_{cluster}$ \Comment{Combine Retrieval Results}

\State \textbf{// Generate Trajectory Pool via Graph}
\State $\mathcal{P} \leftarrow \emptyset$
\For{each $\mathcal{C}_i^M \in \mathcal{R}$}
    \State $\mathcal{N}_k \leftarrow \text{TopK-Neighbors}(\mathcal{C}_i^M, \mathcal{G})$ \Comment{Top-k coding clusters}
    \For{each $\mathcal{C}_j^C \in \mathcal{N}_k$}
        \State Add path $p = (\mathcal{C}_i^M, \mathcal{C}_j^C)$ to $\mathcal{P}$
    \EndFor
\EndFor

\State \textbf{// Rank Paths}
\State $\mathcal{Q} \leftarrow \text{TopM}_{p \in \mathcal{P}}(\text{LLM}_{\text{select}}(p | x_{new}))$

\State \textbf{// Phase 2: Solving via Generalized Knowledge}
\For{each path $p_t = (\mathcal{C}_i^M, \mathcal{C}_j^C) \in \mathcal{Q}$}
    \State \textbf{// Modeling Phase}
    \State $\hat{m}_{raw} \leftarrow \text{LLM}_{\text{gen}}(x_{new} | \mathcal{K}_i^{approach})$
    \State $\hat{m} \leftarrow \text{LLM}_{\text{verify}}(\hat{m}_{raw} | \mathcal{K}_i^{checklist})$
    
    \State \textbf{// Coding Phase}
    \State $\hat{c}_{raw} \leftarrow \text{LLM}_{\text{gen}}(\hat{m} | \mathcal{K}_j^{approach})$
    \State $\hat{c} \leftarrow \text{LLM}_{\text{verify}}(\hat{c}_{raw} | \mathcal{K}_j^{checklist})$
    
    \State \textbf{// Execution \& Repair}
    \State $(result, error) \leftarrow \mathcal{E}(\hat{c})$ \Comment{Execute code}
    
    \If{$error = \text{None}$}
        \Return $\hat{y} = (result, \hat{m}, \hat{c})$
    \Else
        \State \textbf{// Repair Attempt}
        \State $\hat{c}_{fixed} \leftarrow \text{LLM}_{\text{fix}}(\hat{c}, error | \mathcal{K}_j^{pitfall}, \mathcal{K}_j^{checklist})$
        \State $(result, error) \leftarrow \mathcal{E}(\hat{c}_{fixed})$
        \If{$error = \text{None}$}
            \Return $\hat{y} = (result, \hat{m}, \hat{c}_{fixed})$
        \EndIf
    \EndIf
    
    \State \textbf{// Backtrack to next path}
    \State Continue to next $p_{t+1}$ in $\mathcal{Q}$
\EndFor

\Return \textbf{Failure} \Comment{All paths exhausted}
\end{algorithmic}
\end{algorithm*}

\section{Detailed Algorithm Descriptions}
\subsection{Dual-Cluster Memory Construction}
Algorithm~\ref{alg:memory-construction} describes the process of constructing our dual-cluster memory system from historical problem-solution trajectories. 
The algorithm operates in two phases: node-level construction and cluster-level evolution.
In the node-level phase, each historical problem-solution pair is first categorized into three types based on its solving trajectory: Type A (always correct), Type B (recovered), and Type C (persistent failures). 
We then decompose each solution into separate modeling logic $m_i$ and coding implementation $c_i$ components, generate their semantic embeddings $(e_i^m, e_i^c)$, and extract structured knowledge $\Phi_i = \langle \phi_i^{approach}, \phi_i^{checklist}, \phi_i^{pitfall} \rangle$. The approach knowledge captures solution templates, the checklist defines verification criteria, and the pitfall documents common errors.
\par
In the cluster-level phase, we organize nodes into coherent clusters using a hybrid approach combining embedding similarity with LLM-powered semantic verification. 
For each node, its modeling component is assigned to a cluster $\mathcal{C}_j^M$ by retrieving top-$k$ similar clusters and verifying semantic alignment. 
The coding component undergoes identical clustering independently. When a cluster accumulates $N$ new nodes, we trigger knowledge synthesis where an LLM consolidates instance-level knowledge into generalized cluster-level guidelines. 
Finally, the natural associations between modeling and coding clusters form a weighted bipartite graph $\mathcal{G}$, where edge weights indicate co-occurrence frequencies and capture proven compatibility between modeling paradigms and coding strategies.

\subsection{Memory-Augmented Inference}
Algorithm~\ref{alg:memory-inference} presents our inference procedure for solving novel optimization problems using the constructed dual-cluster memory. 
The algorithm implements a systematic Generate-Verify-Repair-Backtrack pipeline guided by cluster knowledge.
\par
The inference begins with a dual-retrieval phase. 
We first encode the new problem into an embedding $e_{new}$ and retrieve the top-$K$ most similar historical instances, capturing fine-grained problem patterns. Complementary to this, we retrieve top-$K$ modeling clusters by comparing $e_{new}$ with cluster centroids, capturing general algorithmic paradigms. 
We combine both retrieval sources and, for each identified modeling cluster, query the bipartite graph to retrieve top-$k$ compatible coding clusters based on edge weights. 
This generates a pool of candidate solution paths $\mathcal{P}$, each representing a complete modeling-to-coding pipeline. An LLM selector then ranks these paths by their alignment with the new problem, forming a prioritized queue $\mathcal{Q}$.

The solving phase executes paths from $\mathcal{Q}$ sequentially. 
For each path, we first generate modeling logic $\hat{m}$ guided by the modeling cluster's approach knowledge and verify it against the checklist. Once validated, we generate executable code $\hat{c}$ using the coding cluster's knowledge and similarly verify it. 
The code is then executed by the solver engine $\mathcal{E}$. On success, we return the solution. 
On failure, we initiate knowledge-guided repair by providing the repair LLM with the error message, cluster-specific pitfalls, and a verification checklist. 
This targeted repair dramatically outperforms generic debugging by grounding fixes in paradigm-specific failure modes. If repair attempts fail within the predefined limit (default: 2 attempts), we backtrack to the next path in $\mathcal{Q}$.

\section{Optimization Solvers and Libraries}
Our framework supports multiple optimization solvers to handle diverse problem types. 
\par
\textbf{Gurobi} is a state-of-the-art free research solver for linear programming (LP), mixed-integer linear programming (MILP), and quadratic programming (QP), providing exact solutions with provable optimality guarantees. Its branch-and-cut algorithm excels at large-scale combinatorial problems such as facility location, scheduling, and resource allocation.
\par
\textbf{PuLP} is an open-source linear programming modeler offering a unified Python interface to various solvers, including CBC, GLPK, and CPLEX. It is particularly suitable for rapid prototyping and straightforward LP/MILP problems. 
\par
\textbf{OR-Tools} is Google's optimization suite featuring constraint programming (CP-SAT) and routing solvers. It handles combinatorial problems with complex logical constraints, such as scheduling with precedence relations, assignment problems, and vehicle routing, supporting both linear and non-linear constraints. 
\par
\textbf{SciPy} provides gradient-based methods (L-BFGS-B, SLSQP) and derivative-free algorithms (Nelder-Mead) for continuous optimization. It is effective for non-linear programming problems, including parameter tuning, curve fitting, and engineering design with non-convex objectives.
\par
\textbf{NetworkX} is a graph analysis library providing efficient implementations of classic algorithms like Dijkstra's shortest path, Ford-Fulkerson maximum flow, and minimum spanning tree. It serves as a specialized solver for network optimization problems with clear graph topology, such as transportation and communication networks.
\par
The bipartite graph in our dual-cluster memory learns associations between modeling paradigms and solver choices through historical co-occurrence patterns. 
For instance, integer linear programs typically pair with Gurobi or OR-Tools clusters, while continuous non-linear problems align with SciPy clusters. 
During inference, the framework automatically selects appropriate solvers by querying this learned graph (Algorithm~\ref{alg:memory-inference}), and supports fallback to alternative solvers if the primary choice fails, ensuring robustness across different settings.

\section{Details of Human Annotators}
For the collection, annotation, and verification of the 500 optimization problems in our dataset, we engage Ph.D. candidates and Master's students with expertise in Operations Research and Applied Mathematics, who are also co-authors of this paper.
All annotators possess strong academic backgrounds, ensuring their qualifications to accurately formulate problems, verify solution correctness, and maintain the technical precision of domain-specific terminology and mathematical notations.
Since the annotators are co-authors involved in this study, no formal external recruitment process or monetary compensation is required, and they are fully informed of the data collection and usage protocols. 
The annotation process focuses exclusively on creating and evaluating optimization problems, mathematical formulations, and solution approaches, without involving the collection of any personally identifying information or exposing annotators to potential risks.
As this research centers on the development and analysis of mathematical optimization content rather than involving external human subjects or sensitive data, it is determined to be exempt from formal institutional review board approval.

\end{document}